\PassOptionsToPackage{dvipsnames,table}{xcolor}
\pdfoutput=1

\documentclass[11pt]{article}

\usepackage[final]{mtsummit25}
\usepackage{copyright}

\usepackage{times}
\usepackage{latexsym}

\usepackage[T1]{fontenc}

\usepackage[utf8]{inputenc}

\usepackage{microtype}

\usepackage{inconsolata}

\usepackage{tabularx}
\usepackage{booktabs}
\usepackage{tipa}
\usepackage{subcaption}
\usepackage{soul}
\setuldepth{t}

\usepackage{multicol}
\usepackage{multirow}
\usepackage{amssymb} 
\usepackage{pifont}
\usepackage{fancybox}
\usepackage{graphicx}

\newcommand{\ap}[1]{{\textcolor{black}{#1}}}

\usepackage{xcolor}
\definecolor{gpt}{HTML}{975e93}
\definecolor{qwen72}{HTML}{f59c29}
\definecolor{qwen32}{HTML}{00a896}

\title{An LLM-as-a-judge Approach for Scalable \\ Gender-Neutral Translation Evaluation}

\author{
 \textbf{Andrea Piergentili\textsuperscript{1,2}},
 \textbf{Beatrice Savoldi\textsuperscript{1}},
 \textbf{Matteo Negri\textsuperscript{1}},
 \textbf{Luisa Bentivogli\textsuperscript{1}},
\\
 \textsuperscript{1}Fondazione Bruno Kessler,
 \textsuperscript{2}University of Trento
\\
 \small{
   \texttt{\{apiergentili,bsavoldi,negri,bentivo\}@fbk.eu}
 }
}

\begin{document}
\maketitle
\begin{abstract}

Gender-neutral translation (GNT) aims to avoid 
expressing the gender of human referents 
when the source text lacks explicit cues about the gender of those referents.
Evaluating GNT automatically is particularly challenging,
with current solutions being limited to monolingual classifiers. 
Such solutions are not ideal because they do not factor in the source sentence and require dedicated data and fine-tuning to scale to new languages.
In this work, we 
address such limitations by investigating
the use of large language models (LLMs) as evaluators of GNT. Specifically, we explore two prompting approaches: one in which LLMs generate 
sentence-level assessments only, and another---akin to a \textit{chain-of-thought} approach---where they first produce detailed phrase-level annotations before 
a sentence-level judgment. 
Through extensive experiments on multiple languages with five models, both open and proprietary, we
show that LLMs can serve as evaluators of GNT. Moreover, we find that prompting for phrase-level annotations before sentence-level assessments consistently improves the accuracy 
of all models, providing a better and more scalable alternative to current 
solutions.\footnote{\ap{Software and data available at \url{https://github.com/hlt-mt/fbk-NEUTR-evAL}}}

\end{abstract}

\section{Introduction}
\label{sec:intro}

Gender-neutral translation (GNT) is the task of translating from one language into another while avoiding gender-specific references in the target text when the source does not provide explicit gender information
\cite{piergentili-etal-2023-gender}.
Consider the examples in Table \ref{tab:gnt-examples}: the source sentence (S)
lacks gender information,
therefore the translation should avoid gendered terms such as `profesor' (masculine) or `profesora' (feminine), and rather use neutral terms like \textit{docente} or the paraphrase \textit{persona que enseña}, thus preserving neutrality.
This serves to
prevent
undesired gender associations in machine translation (MT) outputs, which could result in different types of harm, such as
the reiteration of harmful stereotypes \cite{stanovsky-etal-2019-evaluating,triboulet-bouillon-2023-evaluating}, the unfair representation of gender groups \cite{blodgett-etal-2020-language}, and disparities in 
quality of service \cite{savoldi-etal-2024-harm}.
These issues are especially relevant in 
grammatical gender languages, such as Italian, Spanish, and German, which assign nouns with a grammatical gender and inflect words linked to them accordingly (examples M and F in Table \ref{tab:gnt-examples}).

\setlength{\tabcolsep}{2pt}
\begin{table}
\small
    \centering
    \begin{tabularx}{.48\textwidth}{cX}
    \toprule
       S & Working as a teacher makes me happy \\
       \midrule
       M & Trabajar como \ul{profesor} me hace \ul{contento} \\
       F & Trabajar como \ul{profesora} me hace \ul{contenta} \\
       N\textsubscript{1} & Trabajar como \textit{docente} me hace \textit{feliz} \\
       N\textsubscript{2} & Trabajar como \textit{persona que enseña} me \textit{procura mucha alegría} \textsubscript{[Working as someone who teaches gives me a lot of joy]}\\
    \bottomrule
    \end{tabularx}
    \caption{Examples of gendered and neutral Spanish translations of an English sentence (S) featuring mentions of a human referent and no gender information. Gendered words are underlined, neutral formulations are italicized.}
    \label{tab:gnt-examples}
\end{table}

One of the challenges implied by GNT is \textit{how to evaluate it automatically}, 
thereby enabling fast, cheap, and replicable assessments. 
Indeed, GNT is a complex open natural language generation 
task where individual word choices make the difference between success and failure 
(e.g. en: `as a teacher' → es: `como \ul{una} docente' [F] vs `como docente' [N])
with valid gender-neutralization strategies ranging from pinpointed lexical interventions (example N\textsubscript{1}) to complex and verbose reformulations 
(N\textsubscript{2}) \cite{piergentili-etal-2023-gender}. 
The variability in valid neutralization solutions makes GNT a complicated task to evaluate.
Both traditional and modern MT evaluation metrics struggle to account for these variations, as they are hard to capture at a surface-level \cite{piergentili-etal-2023-hi}, and neutral outputs are systematically penalized by neural metrics \cite{zaranis2024watchingwatchersexposinggender}.
Currently,
the only viable approaches to automatically identify gendered or neutral text rely on dedicated
classifiers \cite{emimic,piergentili-etal-2023-hi}, 
which however do not factor in the source sentence.
Moreover, so far such solutions were only developed for the evaluation of Italian texts, and require
dedicated data and training to scale across languages. The lack of 
easily scalable
evaluation solutions
hinders advancements in GNT research and system development.

To fill this gap,
we look at the \textit{LLM-as-a-Judge} paradigm \cite{gu2025surveyllmasajudge}, where large language models (LLMs) are prompted to perform task-specific evaluations. 
Specifically, we 
ask: \textbf{RQ1) Can we use LLMs to evaluate GNT? RQ2)} 
\textbf{Does 
performing intermediate analytical steps improve the accuracy of LLM-based GNT evaluations?}
To answer these questions, we conduct experiments with five LLMs, evaluating their ability to assess neutrality in Italian, Spanish, and German texts.
We experiment with four evaluation approaches. Two approaches focus on generating sentence-level assessments, while the other two simulate a
\textit{chain-of-thought} \cite{wei2022cot}, prompting LLMs to perform fine-grained phrase-level analysis before providing higher-level judgments.
Overall, we find that LLMs can serve as evaluators of GNT
and that generating phrase-level annotations significantly improves LLMs' accuracy.

\section{Background}

\paragraph{Gender-inclusivity in language technology}

In recent years, the research community's efforts to improve gender fairness in natural language processing (NLP) technologies has grown significantly. With LLMs becoming the state-of-the-art in many NLP tasks \cite{zhao2024surveylargelanguagemodels}, several works have highlighted their shortcomings in language fairness \cite[\textit{inter alia}]{dev2021harmsgenderexclusivitychallenges,lauscher-etal-2022-welcome,hossain-etal-2023-misgendered,waldis-etal-2024-lou}. However, other works identified approaches to improve LLM fairness, both with \cite{bartl-leavy-2024-showgirls} and without \cite{hossain2024misgendermender} fine-tuning.

In MT, while LLMs have been proven not to be immune from gender bias \cite{vanmassenhove2024genderbias,lardelli2024buildingbridges,sant2024powerpromptsevaluatingmitigating}, their in-context learning ability \cite{brown-2020-learners} allowed to mitigate it by controlling the gender in the target sentence \cite{sanchez-etal-2024-gender,lee-etal-2024-fine}.
Moreover, prompting LLMs with more advanced techniques enabled new approaches to gender-inclusive translation, such as using gender-inclusive neopronouns in the target languages \cite{piergentili-etal-2024-enhancing}, or performing GNT \cite{savoldi-etal-2024-prompt}. Here we focus on the latter, for which further progress is still hampered by the lack of automatic evaluation methods.

\paragraph{GNT evaluation}
Evaluating GNT is a complex task due to the variability in valid gender-neutralization strategies, which range from minor lexical substitutions to significant structural reformulations, and are specific to each language, as different languages encode gender differently in their grammar.
Currently, automatic GNT evaluation solutions are based on
fine-tuned BERT-based \cite{devlin-etal-2019-bert} monolingual gender-inclusivity/neutrality classifiers \cite{emimic,savoldi-etal-2024-prompt}. 
This method has significant limitations. 
First,
it does not factor in the source, thus it cannot
assess whether GNT was necessary or appropriate in light of features of the source sentence without dedicated gold labels.
Moreover,
it requires 
task-specific data to fine-tune dedicated models
to scale to other target languages,
with limited flexibility across domains.

To address these limitations,
we look at the emerging \textit{language model-based} approach.

\setlength{\tabcolsep}{1pt}
\begin{table*}[ht]
\small
    \centering
    \begin{tabularx}{\textwidth}{lll}
    \toprule
    Source  & \multicolumn{2}{X}{All this must be carried out in a climate of transparency and regularity so that the citizens do not feel that they are being swindled or sacrificed on the altar of major economic interests.} \\
    Target (\texttt{REF-G}) & \multicolumn{2}{X}{Todo esto se ha de llevar a cabo en un clima de transparencia y de corrección con el fin de que \textbf{\textit{los ciudadanos}} no \textit{se sientan \textbf{estafados}} o \textit{víctimas sacrificadas} en el altar de los grandes intereses económicos.} \\
    \midrule
    {\ding{109}} \textbf{\textsc{Mono-L}}  & \texttt{label:} & \fcolorbox{Lavender!30}{Lavender!30}{\texttt{\color{violet}{GENDERED}}} \\
    \midrule
    \multirow{2}{*}{{\ding{108}} \textbf{\textsc{Mono-P+L}}} & \texttt{phrases:} & \textit{los ciudadanos} \fcolorbox{purple!20}{purple!20}{\texttt{\color{Plum}{M\vphantom{g}}}}, \textit{se sientan estafados} \fcolorbox{purple!20}{purple!20}{\texttt{\color{Plum}{M\vphantom{g}}}}, \textit{víctimas sacrificadas} \fcolorbox{cyan!20}{cyan!20}{\texttt{\color{RoyalBlue}{N\vphantom{g}}}}\vspace{3pt}\\ 
    & \texttt{label:}& \fcolorbox{Lavender!30}{Lavender!30}{\texttt{\color{violet}{GENDERED}}} \\
    \midrule
    $\lozenge$ \textbf{\textsc{Cross-L}} & \texttt{label:} & \fcolorbox{Orchid!20}{Orchid!20}{\texttt{\color{violet}{WRONGLY GENDERED}}} \\
    \midrule
    \multirow{4}{*}{$\blacklozenge$ \textbf{\textsc{Cross-P+L}}} & \texttt{phrases:} &
    \textit{los ciudadanos} \fcolorbox{purple!20}{purple!20}{\texttt{\color{Plum}{M\vphantom{g}}}} \fcolorbox{red!20}{red!20}{\texttt{\color{BrickRed}{wrong\vphantom{N}}}}, \textit{se sientan estafados} \fcolorbox{purple!20}{purple!20}{\texttt{\color{Plum}{M\vphantom{g}}}} \fcolorbox{red!20}{red!20}{\texttt{\color{BrickRed}{wrong\vphantom{N}}}}, \textit{víctimas sacrificadas} \fcolorbox{cyan!20}{cyan!20}{\texttt{\color{RoyalBlue}{N\vphantom{g}}}} \fcolorbox{LimeGreen!20}{LimeGreen!20}{\texttt{\color{teal}{correct\vphantom{gN}}}}\vspace{3pt}\\
    
    & \texttt{label:} & \fcolorbox{Orchid!20}{Orchid!20}{\texttt{\color{violet}WRONGLY GENDERED}} \\
    \bottomrule
     \end{tabularx}
    \caption{Examples of GPT-4o's outputs for each prompt, for a Spanish mGeNTE entry. This is a \texttt{Set-N} entry with a \texttt{REF-G} reference, thus the source includes no gender cue and the target features undue gendered words (in bold). For the \textsc{Mono} prompts ({\ding{109}} and {\ding{108}}) only the target sentence is provided as input, whereas for the \textsc{Cross} prompts ($\lozenge$ and $\blacklozenge$) both the source and target sentences are included.
    \ap{The field \texttt{label} is a sentence-level assessment, whereas \texttt{phrases} is a list of annotations of phrases referred to human beings. Each element of this list includes the piece of text being annotated (in italic), the gender it expresses (\fcolorbox{purple!20}{purple!20}{\texttt{\color{Plum}{M}}}, \fcolorbox{purple!20}{purple!20}{\texttt{\color{Plum}{F}}}, or \fcolorbox{cyan!20}{cyan!20}{\texttt{\color{RoyalBlue}{N}}}), and an 
    assessment
    of whether that gender expression is \fcolorbox{LimeGreen!20}{LimeGreen!20}{\texttt{\color{teal}{correct\vphantom{g}}}} or \fcolorbox{red!20}{red!20}{\texttt{\color{BrickRed}{wrong\vphantom{t}}}} with respect to the information available in the source (only in $\blacklozenge$). Both the \texttt{label} and the list of \texttt{phrases} are generated by the models.}
    } 
    \label{tab:prompt-outputs}
\end{table*}

\paragraph{LLM-as-a-Judge}
Recently, LLMs have been successfully employed as evaluators of 
natural-language generation
tasks \cite{wang2023chatgptgoodnlgevaluator,liu2023gevalnlgevaluationusing,bavaresco2024llmsinsteadhumanjudges} including MT, where proprietary LLMs have been employed as state-of-the-art MT quality evaluators \cite{kocmi-federmann-2023-large,leiter-eger-2024-prexme} without the need for dedicated fine-tuning data. 
Several works used LLMs to provide insights into fine-grained aspects, such as fluency, accuracy, and style \cite{fu-etal-2024-gptscore,lu-etal-2024-error}. 
Moreover, LLMs have successfully been employed to generate the error annotations required for Multidimensional Quality Metrics 
assessments \cite{fernandes-etal-2023-devil,kocmi-federmann-2023-gemba,feng2024mmad,zouhar2025esaai}, an evaluation paradigm designed for human evaluators, which requires pinpointed analyses and attention to context.
Furthermore, and 
more related
to our work, LLMs have also been found to be accurate evaluators of masculine/feminine references to human beings in monolingual contexts \cite{derner2024leveragingllms}.


Here, we investigate whether LLMs' ability to 
generate fine-grained assessments can be leveraged to build a GNT evaluation method scalable across languages without the need for dedicated fine-tuning data.

\setlength{\tabcolsep}{2pt}
\begin{table*}[ht]
\small
    \centering
    \begin{tabularx}{\textwidth}{ccX}
    \toprule
        \textbf{\texttt{Set-G}} & \texttt{SRC} (F) &  \textbf{Madam} President, I should like to thank \textbf{Mrs} Oostlander for \textbf{her} sterling contribution as delegate.  \\
    \midrule
        de & \fcolorbox{YellowGreen!40}{YellowGreen!40}{\texttt{REF-G}} & \textbf{Frau Präsidentin}! Ich möchte \textbf{der Kollegin} Oostlander für \textbf{ihre} verdienstvolle Arbeit als Delegierte danken. \\
        de & \texttt{REF-N} & Geehrtes Präsidium! Ich möchte dem Kollegiumsmitglied Oostlander für seine verdienstvolle Arbeit als Delegierte danken. \\
        es & \fcolorbox{YellowGreen!40}{YellowGreen!40}{\texttt{REF-G}} & \textbf{Señora Presidenta}, quiero agradecer a \textbf{la Sra.} Oostlander sus valiosos esfuerzos como \textbf{delegada}. \\
        es & \texttt{REF-N} & Con la venia de la Presidencia, quiero agradecer a su Señoría Oostlander sus valiosos esfuerzos como integrante de la delegación.  \\
        it & \fcolorbox{YellowGreen!40}{YellowGreen!40}{\texttt{REF-G}} & \textbf{Signora} Presidente, ringrazio \textbf{la} onorevole Oostlander per il lavoro meritorio che ha assolto come \textbf{delegata}. \\
        it & \texttt{REF-N} & Gentile Presidente, ringrazio l'onorevole Oostlander per il lavoro meritorio che ha assolto come membro della delegazione. \\
    \toprule
        \textbf{\texttt{Set-N}} & \texttt{SRC} & There are no better guardians of the Treaties than the European citizens. \\
    \midrule
        de & \texttt{REF-G} & Niemand eignet sich als \textbf{Hüter} der Verträge besser als die europäischen \textbf{Bürger}. \\
        de & \fcolorbox{YellowGreen!40}{YellowGreen!40}{\texttt{REF-N}} & Niemand eignet sich zum Hüten der Verträge besser als die europäische Bevölkerung. \\
        es & \texttt{REF-G} & No hay mejores \textbf{custodios} de los Tratados que \textbf{los ciudadanos europeos}. \\
        es & \fcolorbox{YellowGreen!40}{YellowGreen!40}{\texttt{REF-N}} & No hay mejores vigilantes de los Tratados que la ciudadanía europea. \\
        it & \texttt{REF-G} & \textbf{I} migliori \textbf{guardiani} dei Trattati sono \textbf{gli stessi cittadini europei}.   \\
        it & \fcolorbox{YellowGreen!40}{YellowGreen!40}{\texttt{REF-N}} & Le popolazioni residenti sul suolo europeo sono le migliori custodi dei Trattati.  \\
    \bottomrule
    \end{tabularx}
    \caption{Examples of mGeNTE entries from \texttt{Set-G} and \texttt{Set-N}, with both \texttt{REF-G} and \texttt{REF-N}, and parallel across the three target languages. Gender cues in the source and gendered words in the references are in bold. The 
    matching
    reference for the entry is highlighted.}
    \label{tab:mgente_examples}
\end{table*}

\section{GNT evaluation prompts}
\label{sec:prompts}
To explore the LLM-as-a-Judge approach to GNT evaluation and investigate whether prompting for intermediate analytical steps leads to higher evaluation accuracy, we
experiment with four prompts targeting different approaches to evaluation. 
We design prompts dedicated to the evaluation of the target language text only (\textsc{Mono}) and the source sentence paired with the translation (\textsc{Cross}).
With the \textsc{Mono} prompts we attempt to replicate the evaluation enabled by the gender-neutrality classifier introduced in \citet{piergentili-etal-2023-hi} on new languages, without generating dedicated data and fine-tuning new models.
\textsc{Mono} prompts can be used \textit{as is} to evaluate intra-lingual neutral rewriting tasks \cite{vanmassenhove-etal-2021-neutral,veloso-etal-2023-rewriting,frenda-etal-2024-gfg}. However, for GNT evaluation, they still require gold labels specifying whether the source sentence should be translated neutrally, as with classifier-based methods. The \textsc{Cross} prompts instead task the models not just with classifying the target text as gendered or neutral, but also to determine if the target's gender correctly aligns with the source sentence. This allows for GNT evaluation in realistic scenarios, i.e. outside of the benchmarking enabled by gold source sentence labels.

For 
both the \textsc{Mono} and \textsc{Cross} approaches
we experiment with one 
prompt that requires
the model to generate a sentence-level label only (\texttt{L}) and another where the model must generate phrase-level annotations first and then provide a sentence-level label (\texttt{P+L}).
Examples of the output of each prompt are available in Table \ref{tab:prompt-outputs}. Complete instructions and further details 
are provided in Appendix \ref{app:prompts}.

\paragraph{{\ding{109}} \textsc{Mono-L}} We provide the model with the target sentence and instruct it to classify the sentence as 
\fcolorbox{Lavender!20}{Lavender!20}{\texttt{\color{violet}{GENDERED}}}
if at least one masculine or feminine reference to human beings is found, or as 
\fcolorbox{Turquoise!20}{Turquoise!20}{\texttt{\color{Blue}{NEUTRAL}}}
if the whole sentence is gender-neutral. 
This 
prompt
does not imply any 
intermediate annotation, only 
requesting
one sentence-level label. 

\paragraph{{\ding{108}} \textsc{Mono-P+L}} 
The model is instructed to first generate annotations for all phrases that refer to human beings in the target sentence. For each phrase, the model must also provide a label indicating its semantic gender: 
\fcolorbox{purple!20}{purple!20}{\texttt{\color{Plum}{M}}}
(masculine), 
\fcolorbox{purple!20}{purple!20}{\texttt{\color{Plum}{F}}}
(feminine), or 
\fcolorbox{cyan!20}{cyan!20}{\texttt{\color{RoyalBlue}{N}}}
(neutral). Finally, the model must provide the same sentence-level label as in \textsc{Mono-L},
based on the same principle: if one or more of the annotated phrases is gendered, the sentence label should be 
\fcolorbox{Lavender!20}{Lavender!20}{\texttt{\color{violet}{GENDERED}}};
otherwise, it should be 
\fcolorbox{Turquoise!20}{Turquoise!20}{\texttt{\color{Blue}{NEUTRAL}}}.
This prompt 
introduces intermediate annotations, which are expected to inform the models' choice of the final sentence-level label.

\paragraph{$\lozenge$ \textsc{Cross-L}} We provide the model with both the source and target sentences, instructing it to classify the target as 
\fcolorbox{Turquoise!20}{Turquoise!20}{\texttt{\color{Blue}{NEUTRAL}}}
if fully gender-neutral,  
\fcolorbox{Apricot!20}{Apricot!20}{\texttt{\color{violet}{CORRECTLY GENDERED}}}
if it accurately reflects gender information from the source, or 
\fcolorbox{Orchid!20}{Orchid!20}{\texttt{\color{violet}{WRONGLY GENDERED}}}
if the target's gender does not match the source or the target adds gender information when the source lacks it.
We do not distinguish between \textit{correct} and \textit{incorrect} \texttt{NEUTRAL} 
translations. While using gendered language when gender is unspecified in the source is undesirable (i.e. \fcolorbox{Orchid!20}{Orchid!20}{\texttt{\color{violet}{WRONGLY GENDERED}}}), neutral translations---though not always necessary---merely avoid gender marking and therefore cannot be considered wrong by definition.\footnote{We note that there are instances
where gender is essential to the meaning of a sentence and should then be preserved in translation.
For example, to refer to specific groups, as in \textit{`men tend to suffer from heart attacks at higher rates'}). Accounting for this aspect in the evaluation requires finer-grained analyses factoring in translation adequacy as well.
As such instances represent less than 3\% of our test data, we retain them in our experiments and leave this analysis to future work.}

\paragraph{$\blacklozenge$ \textsc{Cross-P+L}} We instruct the model to generate the same annotations as \textsc{Mono-P+L}, with the addition of an assessment of whether each phrase's gender is 
\fcolorbox{LimeGreen!20}{LimeGreen!20}{\texttt{\color{teal}{correct\vphantom{g}}}} or \fcolorbox{red!20}{red!20}{\texttt{\color{BrickRed}{wrong\vphantom{t}}}}
with respect to the source. Finally, the model must provide the same sentence-level label as in \textsc{Cross-L}. Similarly to \textsc{Mono-P+L}, this prompt introduces the intermediate phrase annotations that the model is expected to leverage to provide more accurate labels.

\section{Experimental settings}
\label{sec:experimental-settings}
We experiment with LLM-based evaluation of GNT from English into three target languages---Italian, Spanish, and German---
in two scenarios: 
\begin{itemize}
\item \textbf{\textit{Target-only}}, where LLMs only 
evaluate
the target language text. 
In this scenario, models are tasked with assessing whether the text contains any gendered mention of human beings and label it
\fcolorbox{Lavender!20}{Lavender!20}{\texttt{\color{violet}{GENDERED}}}, or no such mention and label it 
\fcolorbox{Turquoise!20}{Turquoise!20}{\texttt{\color{Blue}{NEUTRAL}}}.

\item\textbf{\textit{Source-target}}, where LLMs receive both the source sentence and the target language translation as input. 
Here, the models must assess whether the target language text is 
\fcolorbox{Turquoise!20}{Turquoise!20}{\texttt{\color{Blue}{NEUTRAL}}},
\fcolorbox{Apricot!20}{Apricot!20}{\texttt{\color{violet}{CORRECTLY GENDERED}}}, or 
\fcolorbox{Orchid!20}{Orchid!20}{\texttt{\color{violet}{WRONGLY GENDERED}}} with respect to the information available in the source. 
\end{itemize}

\subsection{Test data and evaluation metrics}
\label{sec:test-data}


We conduct our experiments on mGeNTE \cite{savoldi2025mgente}, 
a multilingual test set for GNT. Available for en-it/es/de, for each language pair it comprises
1,500 parallel sentences, 
evenly divided in two subsets (see Table \ref{tab:mgente_examples}): 
\texttt{Set-G} entries feature words in the source that provide information about the gender of human referents
(e.g., \textit{Madam}, \textit{Mrs}, and \textit{her} in the \texttt{Set-G} example in Table \ref{tab:mgente_examples})
whereas \texttt{Set-N} entries do not.
The \texttt{Set-G} sentences are further split into masculine-only and feminine-only, and labeled \texttt{M} and \texttt{F} respectively. 

 
We use mGeNTE as the test set to validate our evaluation approaches because it provides dedicated human-made translations and gold labels for GNT. Although evaluating human-made translations is not fully representative of realistic conditions, this dataset remains the only multilingual resource available with GNT-specific gold labels. To further explore automatic GNT evaluation in realistic conditions, we also experiment with model-generated translations of a subset of mGeNTE. 

mGeNTE references and the automatic translations we use in our experiments are described in sections \ref{sec:mgente-references} and \ref{sec:mgente-outputs} respectively. Statistics on our experimental data are reported in Table \ref{tab:data-stats}.

\setlength{\tabcolsep}{2pt}
\begin{table}
\small
    \centering
    \begin{tabular}{lcccc}
    \toprule
    \textsc{Set} & \textsc{Split} & \texttt{GENDERED} & \texttt{NEUTRAL} & TOTAL \\
    \midrule
        \textbf{mGeNTE references} & \texttt{Set-G} & 750 & 750 & 1,500 \\
         (x3: en-it/de/es) & \texttt{Set-N} & 750 & 750 & 1,500 \\
    \midrule
        \textbf{Automatic GNTs} & \multirow{2}{*}{\texttt{Set-N}}  & \multirow{2}{*}{340} & \multirow{2}{*}{740} & \multirow{2}{*}{1,080} \\
        (en-it only) & & &\\
    \bottomrule
    \end{tabular}
    \caption{Statistics about the test data. 
    mGeNTE values
    are referred
    to each target language, whereas the automatic GNTs 
    are available only for
    en-it.}
    \label{tab:data-stats}
\end{table}

\subsubsection{mGeNTE references}
\label{sec:mgente-references}
To run experiments on Italian, Spanish, and German texts, we use the reference translations in mGeNTE.
Each source sentence in the corpus corresponds to two reference translations produced by professionals: a gendered reference (\texttt{REF-G}), considered ideal for \texttt{Set-G} but incorrect for \texttt{Set-N}, and a gender-neutral reference (\texttt{REF-N}), correct for \texttt{Set-N} and not ideal for \texttt{Set-G}.
We use both \texttt{REF-G} and \texttt{REF-N} in isolation as input in our 
\textit{target-only}
scenario, and we pair them with the source sentence in the 
\textit{source-target}
scenario.

To evaluate on this data set, we
compute sentence-level label accuracies by matching models' predictions against the true labels in mGeNTE.
We use the data split labels (\texttt{Set-G} and \texttt{Set-N}) in combination with the reference labels to determine the true labels for each scenario and data split. 
For the 
\textit{target-only}
scenario, we map \texttt{REF-G} and \texttt{REF-N} to 
\fcolorbox{Lavender!20}{Lavender!20}{\texttt{\color{violet}{GENDERED}}} and 
\fcolorbox{Turquoise!20}{Turquoise!20}{\texttt{\color{Blue}{NEUTRAL}}} respectively. For the 
\textit{source-target}
scenario, \texttt{REF-G} is further categorized as 
\fcolorbox{Apricot!20}{Apricot!20}{\texttt{\color{violet}{CORRECTLY GENDERED}}}
for \texttt{Set-G} entries and 
\fcolorbox{Orchid!20}{Orchid!20}{\texttt{\color{violet}{WRONGLY GENDERED}}}
for \texttt{Set-N} entries.

\subsubsection{Automatic GNTs}
\label{sec:mgente-outputs}
We also experiment on a more realistic evaluation scenario, where evaluator LLMs are tasked with assessing 
automatic gender-neutralizations instead of human references,
using
a set of automatic translations of mGeNTE en-it sentences taken from \texttt{Set-N} \citep{savoldi-etal-2024-prompt}.\footnote{\url{https://mt.fbk.eu/gente/}} The translations were produced by GPT-4\footnote{Model \texttt{gpt-4-0613}} \cite{openai2024gpt4technicalreport} 
and manually evaluated by human experts, who provided gold labels about the neutrality of the outputs.\footnote{The outputs were originally divided into \textit{neutral}, \textit{partially neutral}, and \textit{gendered}. Here, we adjusted this tripartition to our label system by merging the \textit{partially neutral} category into the 
\fcolorbox{Lavender!20}{Lavender!20}{\texttt{\color{violet}{GENDERED}}}
label, in line with the classifier's binary label system.}

As the classes in this dataset 
are unbalanced (see Table \ref{tab:data-stats}),
to assess the performance of evaluator LLMs
we compute precision and recall scores rather than a simple label accuracy in this case. 
Since all the source sentences from this data set originally belonged to \texttt{Set-N}, we consider 
\fcolorbox{Turquoise!20}{Turquoise!20}{\texttt{\color{Blue}{NEUTRAL}}} 
the positive class. Moreover, since this set of 
automatic GNTs
does not include \texttt{Set-G} entries, and consequently
one of the three labels from the \textit{source-target} scenario 
(\fcolorbox{Apricot!20}{Apricot!20}{\texttt{\color{violet}{CORRECTLY GENDERED}}}) 
is not represented within it, we only use this data set in the \textit{target-only} scenario.


\setlength{\tabcolsep}{4pt}
\begin{table}
\small
    \centering
    \begin{tabular}{lccc}
    \toprule
     \textbf{Model} & \textbf{en-de} & \textbf{en-es} & \textbf{en-it} \\
     \midrule
     Tower 13B &  0.4407 & 0.4610 & 0.4587 \\
     \midrule
     GPT-4o  &  \underline{0.4635}  & \underline{0.4720}  &  \underline{0.4730}  \\
     Qwen 32B &  \underline{0.4485}  &  0.4608  &  \underline{0.4601}  \\
     Qwen 72B &   \underline{0.4533}  &  \underline{0.4647}  &  \underline{0.4646}  \\
     Mistral Small &   \underline{0.4552}  &  \underline{0.4623}  &  \underline{0.4623} \\
     DS Qwen 32B &  0.4365  &  0.4559  &  0.4517 \\
    \bottomrule
    \end{tabular}
    \caption{COMET scores of all models' MT outputs on FLORES+. Instances where one of the models outperform Tower 13B are underlined.}
    \label{tab:results-mt}
\end{table}

\subsection{Prompting details}
\label{sec:prompting-details}


For each of the prompts described in section \ref{sec:prompts},
we use the same
8 task exemplars to elicit LLMs' in-context learning \cite{brown-2020-learners,min_rethinking_2022}. 
These exemplars were selected from mGeNTE entries parallel across the three languages, 
and were balanced
across
the \texttt{Set-G/N}, \texttt{REF-G/N}, and gender combinations. The entries used as exemplars were excluded from the test data in the experiments.

\begin{figure*}[h]
    \centering
    \includegraphics[width=1\linewidth]{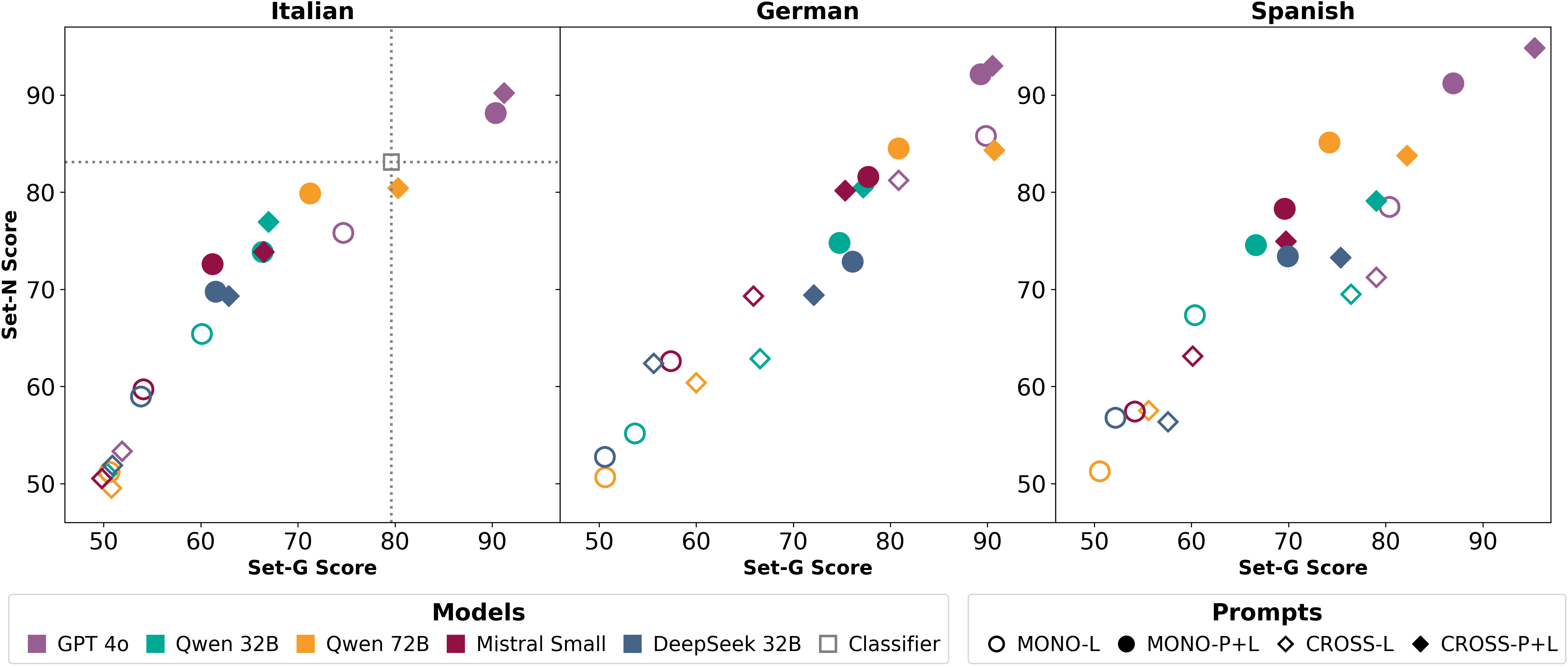}
    \caption{Accuracy of all models in \textbf{\textit{target-only}} GNT evaluation experiments on \textbf{mGeNTE references}. The Italian experiments include the performance of the gender-neutrality classifier, which is not available for other languages.}
    \label{fig:mono-results}
\end{figure*}

We constrain models' generations to adhere to specific JSON schemas via structured generation \cite{willard2023efficient}, which, at each generation step, restricts the model's vocabulary to the tokens allowed at that step by the schema, masking out the invalid ones. This ensures that all models' outputs adhere to the same formats without the need for post-processing or parsing within open ended generations.

\subsection{Models}
\label{sec:models}
We experiment with open models of different families and sizes: Qwen 2.5 32B and 72B \cite{qwen2.5}, Mistral Small 3 24B,\footnote{\url{https://mistral.ai/en/news/mistral-small-3}} DeepSeek-R1-Distill-Qwen-32B \cite{deepseekai2025r1}.\footnote{We performed a first selection of models that perform best on instruction following tasks on Open LLM Leaderboard \cite{open-llm-leaderboard-v2}, then further selected the models that performed best in preliminary experiments.} We also include GPT-4o\footnote{Model \texttt{gpt-4o-2024-08-06}} \cite{openai2024gpt4ocard} as representative of the closed, commercial models.
All models are fine-tuned for instruction following \cite{ouyang2022instructgpt,chung2024scaling}.

To ensure that the models we selected perform well on the target languages we include in our experiments, we assess all models' performance on generic translation into Italian, Spanish, and German
using FLORES+ \cite{nllb-24}. Table \ref{tab:results-mt} reports their 
COMET\footnote{Model \texttt{Unbabel/wmt22-cometkiwi-da} \cite{rei-etal-2022-cometkiwi}} \cite{rei2020comet} scores, which 
measure 
how well model outputs represent the source sentence meaning.
As a 
baseline
we report the performance of Tower 13B Instruct \cite{tower_llm_2024}, a state-of-the-art open LLM fine-tuned for MT tasks. All models were prompted to perform MT with default settings and three shots randomly selected from the dev split of FLORES+. 
%
The results show that all models perform well in all target languages compared to Tower 13B, indicating consistently strong performance across the languages evaluated.

\section{Results and discussion}
\label{sec:results}

\begin{figure}
    \centering
    \includegraphics[width=.99\linewidth]{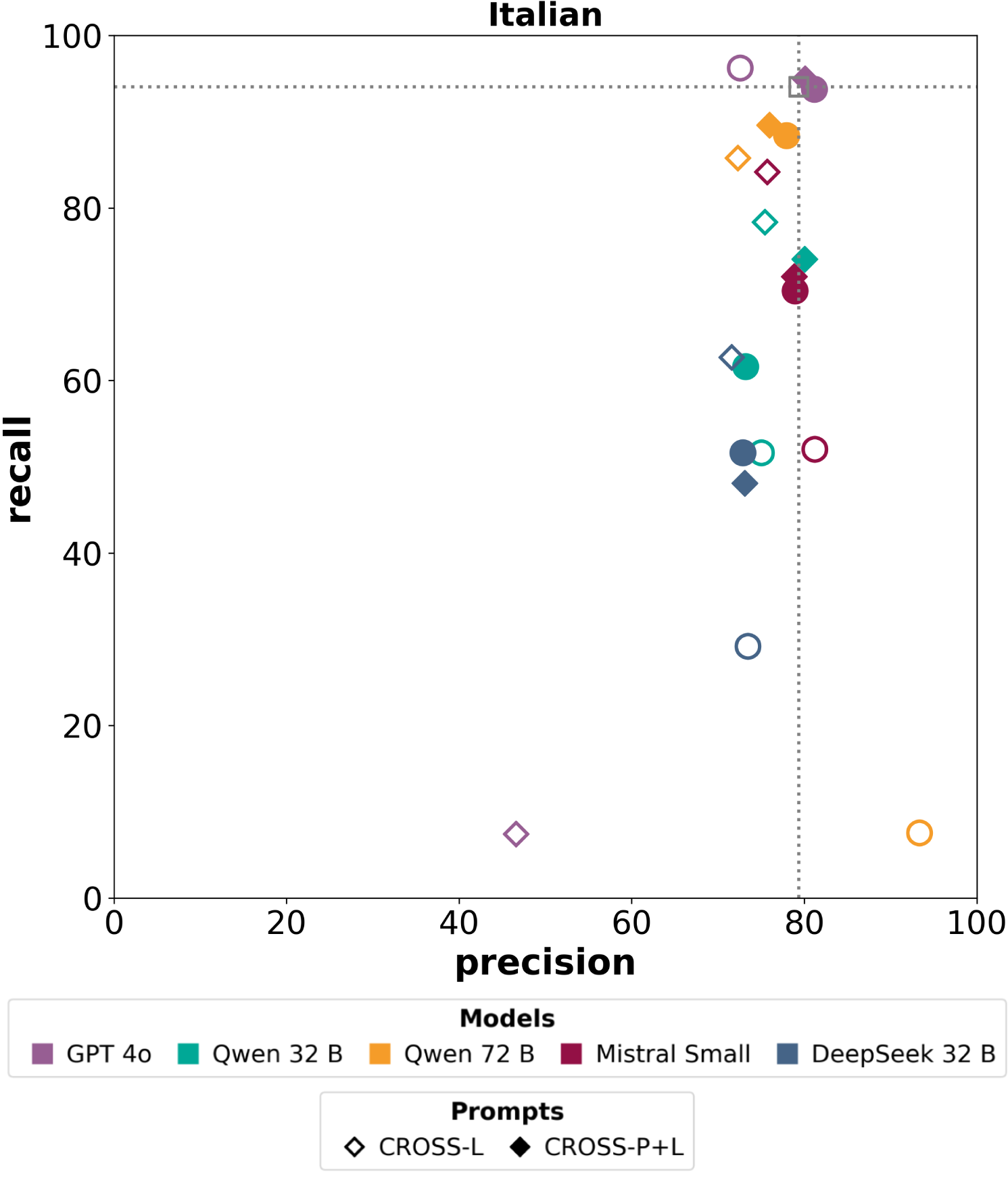}
    \caption{Precision and recall scores of all models in \textit{\textbf{target-only}} GNT evaluation 
    of \textbf{automatic GNTs}.}
    \label{fig:eacl-mono}
\end{figure}

\begin{figure*}[ht]
    \centering
    \includegraphics[width=1\linewidth]{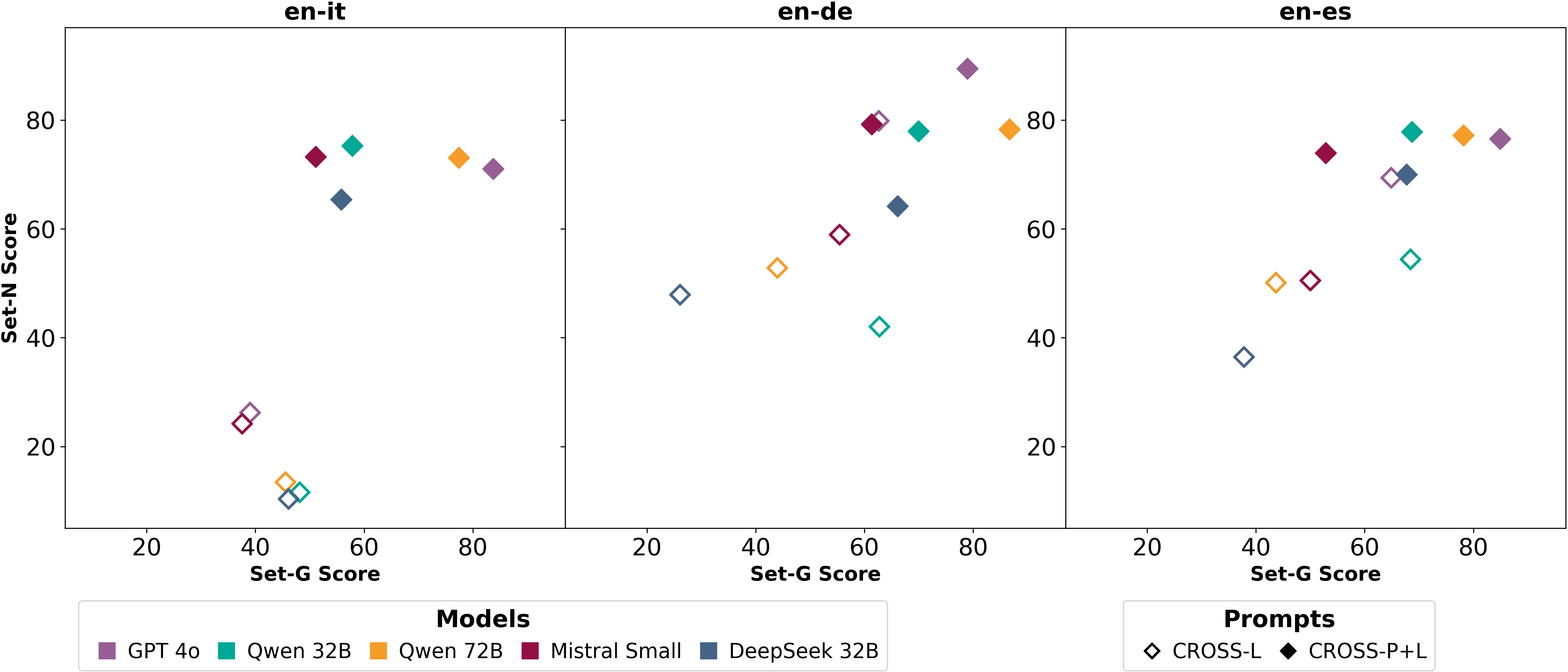}
    \caption{Accuracy of all models in \textbf{\textit{source-target}} GNT evaluation experiments on \textbf{mGeNTE source-reference} pairs. Note that the axes here encompass a wider range of values compared to the \textit{target-only} chart.}
    \label{fig:cross-results}
\end{figure*}

We report the results of our experiments in \textit{target-only} and \textit{source-target} GNT evaluation experiments on mGeNTE references in Figures \ref{fig:mono-results} and \ref{fig:cross-results} respectively. Results on 
automatic GNTs
are reported in Figure \ref{fig:eacl-mono}.
In the \textit{target-only} charts we include the performance of the gender-neutrality classifier\footnote{\url{https://huggingface.co/FBK-MT/GeNTE-evaluator}} on the test data as a baseline for the Italian experiments. The detailed results of all evaluation experiments are reported in Appendix \ref{app:results}, along with additional discussions.

To make the performance of the \textsc{Mono} and \textsc{Cross} prompts comparable in the \textit{target-only} scenario, we 
count
the labels 
\fcolorbox{Apricot!20}{Apricot!20}{\texttt{\color{violet}{CORRECTLY GENDERED}}}
and
\fcolorbox{Orchid!20}{Orchid!20}{\texttt{\color{violet}{WRONGLY GENDERED}}}
as correct matches of 
\fcolorbox{Lavender!20}{Lavender!20}{\texttt{\color{violet}{GENDERED}}}.

\subsection{\textit{Target-only} evaluation}
\label{sec:results-mono}

\paragraph{Results on mGeNTE references}
Looking at the \textit{target-only} results we note that
GPT-4o is consistently the best overall performer, and the only model outperforming the gender-neutrality classifier in the Italian scenario (90.72\% vs 81.37\% overall accuracy, see Table \ref{tab:mono-results-it}). Among the open models, Qwen 2.5 72B performs best and comes close to the classifier's performance with the \textsc{Cross-P+L} prompt (\textcolor{qwen72}{$\blacklozenge$}).
We note that all models perform better in the Spanish and German experiments, with GPT-4o reaching 95.08\% overall accuracy in the latter (\textcolor{gpt}{$\blacklozenge$}) and Qwen 2.5 72B (\textcolor{qwen72}{\ding{108}}, \textcolor{qwen72}{$\blacklozenge$}) and 32B (\textcolor{qwen32}{$\blacklozenge$}) showing solid performance in both languages. 
To answer \textbf{RQ1}, the \textit{target-only} results indicate that \textbf{LLMs can serve as evaluators of gender-neutrality in multiple languages with good accuracy}.

To answer \textbf{RQ2}, we compare
the performance of the four prompting strategies 
and
notice that \textbf{the \textsc{P+L} 
prompts
(\ding{108} and $\blacklozenge$) produce more accurate results} than the label-only 
prompts
(\ding{109} and $\lozenge$) across all languages.
Furthermore, the richer annotation 
prompt
\textsc{CROSS-P+L} ($\blacklozenge$) generally outperforms the others. 
We conclude
that \textbf{guiding models to generate intermediate finer-grained annotations improves the accuracy of the sentence-level assessments}.


\paragraph{Results on automatic GNTs}
Results on the \textit{target-only} experiments on automatic GNTs (Figure \ref{fig:eacl-mono}), confirm the findings discussed above. We find analogous rankings, with only GPT-4o slightly outperforming the classifier (\textcolor{gpt}{\ding{108}}, \textcolor{gpt}{{$\blacklozenge$}}) and Qwen 2.5 72B being the best open model. Comparing prompt strategies, here we see the \textsc{P+L} 
prompts
outperforming the others---though only for the best models, with no coherent trend for the others.

Specific to these results, looking at precision and recall we note that most of the model/prompt combinations produce similar precision values, meaning that they show similar abilities in correctly labeling gendered sentences (few false positives). It is the recall score that ultimately makes the difference in performance, i.e. their ability to correctly label neutral sentences.

\subsection{\textit{Source-target} evaluation results}
\label{sec:results-cross}

The results of the \textit{source-target} evaluation experiments (Figure \ref{fig:cross-results}) support our previous findings. 
First, we observe that, with dedicated prompting, \textbf{LLMs can serve as multilingual evaluators of GNT with solid accuracy. This enables the evaluation of GNT 
in absence of gender information about the source sentence}.
Second, we confirm that \textbf{phrase annotation ($\blacklozenge$) consistently improves evaluation accuracy across all models}. 

Moreover, similarly to the \textit{target-only} scenario, GPT-4o outperforms the open models in this setting as well. All models generally exhibit better performance on Spanish and German rather than Italian in cross-lingual evaluation here too.
Overall, scores are generally lower than in the \textit{target-only} setting. This is likely due to the further distinction of the 
\fcolorbox{Lavender!20}{Lavender!20}{\texttt{\color{violet}{GENDERED}}} label into 
\fcolorbox{Apricot!20}{Apricot!20}{\texttt{\color{violet}{CORRECTLY}}}
and 
\fcolorbox{Orchid!20}{Orchid!20}{\texttt{\color{violet}{WRONGLY GENDERED}}},
which increases task complexity, and to the added challenge of incorporating the source sentence. 
As found by \citet{huang-etal-2024-lost}, 
LLMs generally perform better in MT evaluation when provided with a reference translation rather than the source. Our results reflect this trend, suggesting that despite their strong translation capabilities, \textbf{LLMs still have limited ability to leverage cross-lingual information for evaluation}. Additionally, model performance gaps are narrower in this scenario,
indicating that even the models that perform best in \textit{target-only} evaluation are not immune to these limitations.

\section{Conclusions}
\label{sec:conclusion}
We investigated several LLMs' ability in performing GNT evaluation 
across three target languages---Italian, German, and Spanish---comparing their performance against the previously available solutions, namely classifier-based approaches.
We experimented with two prompting approaches and constrained LLMs to generate sentence-level labels only in one case and phrase-level annotation as well as sentence-level labels in the other, with the latter being akin to a 
\textit{chain-of-thought} approach.
Our experimental results show 
that guiding the models to generate fine-grained annotations before providing a higher level assessment significantly improves their accuracy.
However, while in 
target language
evaluation some of the models reach almost perfect accuracy, 
assessing neutrality with reference to the source sentence
emerges as a harder task for LLM evaluators, in line with findings from the literature.
Overall, our LLM-based approach outperforms existing solutions, and provides a scalable method for automatic GNT evaluation that generalizes effectively across languages, without requiring additional task-specific data.

\section{Limitations}
Naturally, our work comes with some limitations. First, while the explored evaluation approaches address some of the limitations of previous solutions, they are still confined to discrete sentence-level labels. These make our approaches unable to distinguish degrees of success or failure in using the appropriate gender expression in relation to specific human referents. For instance, this means that we are not able to assess whether a non-neutral output includes only one gendered mention of a human entity or multiple ones, thus we cannot perform more nuanced analyses or rankings of different outputs or systems.
Second, our approaches do not factor in an important aspect of GNT: the acceptability of neutral text \cite{savoldi-etal-2024-prompt}. Acceptability is a complex aspect, determined by the adequacy of the target text with respect to the source sentence meaning and its fluency in the target language. Developing evaluation systems that can account for acceptability in the evaluation calls for dedicated research work, and the validation of such systems requires fine-grained human reference annotations that are currently not available. 
For similar reasons, in our analyses we only focused on the sentence-level annotations generated by the models, and leaving aside the phrase-level annotations generated with the \textsc{P+L} prompts. Since the mGeNTE corpus does not include 
gold
annotations of phrases that refer to human beings 
we only measured 
sentence-level accuracy and could not evaluate the phrase-level annotations generated by the models.

While we are interested in exploring all the aspects mentioned above in the future, with this work we focused on tackling limitations of previously available solutions, enabling and improving the evaluation of GNT across new languages, and doing so with an easy to replicate method, so as to foster the development and research of GNT in further languages as well. 


\section{Bias statement}
This paper addresses representational and allocational harms as defined by \citet{blodgett-etal-2020-language} arising from gender biases in automatic translation, particularly when translation systems unnecessarily default to gender-specific terms, perpetuating harmful stereotypes or excluding non-binary identities. Our approach leverages LLMs for scalable, automatic evaluation of GNT. We assume GNT is desirable when the source text does not specify gender explicitly, aiming to prevent unfair gendered expressions. However, we acknowledge that gender-neutral language is only one of the approaches to inclusive language \cite{lardelli-gromann-2023-gender}, and is not necessarily perceived as inclusive by all speakers \cite{spinelli2023neutral}. We also acknowledge potential biases inherent to LLMs from their training data, possibly affecting evaluation outcomes across languages and cultural contexts.



\section*{Acknowledgments}
\ap{This paper has received funding from the European Union’s Horizon research and innovation programme under grant agreement No 101135798, project Meetween (My Personal AI Mediator for Virtual MEETings BetWEEN People).}
\ap{We also acknowledge the support of the PNRR project FAIR - Future AI Research (PE00000013), under the NRRP MUR program funded by the NextGenerationEU.}
\ap{Finally, we acknowledge the CINECA award under the ISCRA initiative (AGeNTE), for the availability of high-performance computing resources and support.}

\bibliography{custom}

\appendix

\section{Prompting details}
\label{app:prompts}

We include verbalized instructions and attribute descriptions in English to facilitate the models' understanding of the task \cite{dey2024betteraskenglishevaluation}. We report the system instruction of prompts \textsc{Mono-L}, \textsc{Mono-P+L}, \textsc{Cross-L}, and \textsc{Cross-P+L} in Tables \ref{tab:prompt-Mono-L}, \ref{tab:prompt-Mono-P+L}, \ref{tab:prompt-Cross-L}, and \ref{tab:prompt-Cross-P+L} respectively.


\begin{table*}
\small
    \centering
    \begin{tabularx}{\textwidth}{|X|}
    \hline
     You are a language expert specializing in evaluating gender neutrality in Italian texts. Your task is to assess each provided sentence and determine whether it is gendered or neutral.\\
     \\Guidelines:\\
     \\1. Identify relevant phrases: carefully analyze the Italian sentence and focus on all phrases that refer to human beings or groups of human beings, including:\\
     - Noun phrases (e.g., "un’ottima oratrice", "la cittadinanza"),\\
     - Verb phrases (e.g., "è molto felice", "ho purtroppo dovuto"),\\
     - Adjective phrases (e.g., "felicemente sposato", "molto competente").\\
     \\
     2. Evaluate gender information: consider only the social gender conveyed by the phrases, not grammatical gender. For example:
     \\-  Phrases like "un oratore", "è molto contento", "tutti i colleghi", and "i cittadini" are masculine;
     \\-  Phrases like "un’oratrice", "è molto contenta", "tutte le colleghe", and "le cittadine" are feminine;
     \\- Phrases like "una persona che parla in pubblico", "è molto felice", "tutte le persone con cui lavoro", and "la cittadinanza" do not express social gender, therefore they must be considered neutral.\\
     
     \\3. Assign a label:
     \\- If all references to human beings are gender-neutral, label the sentence as "NEUTRAL".
     \\- If one or more expressions convey a specific masculine or feminine gender, label the sentence as "GENDERED".
   \\
   \hline
   
    \end{tabularx}
    \caption{System message for prompt \textsc{Mono-L} (Italian).}
    \label{tab:prompt-Mono-L}
\end{table*}

\begin{table*}
\small
    \centering
    \begin{tabularx}{\textwidth}{|X|}
    \hline
     You are a language expert specializing in evaluating gender neutrality in German texts. Your task is to extract target German phrases that refer to human beings and determine whether each phrase is masculine, feminine, or neutral. Based on the phrases, assess whether the sentence is gendered or neutral.\\

   \\ Guidelines:\\
    \\1. Identify relevant phrases: carefully analyze the German sentence and focus on all phrases that refer to human beings or groups of human beings (e.g., "eine ausgezeichnete Rednerin", "die Bürgerschaft", "Sie").\\
    
    \\2. Evaluate gender information: consider only the social gender conveyed by the phrases, not grammatical gender, and assign a label to each phrase [M/F/N]. For example:
    \\- Phrases like "Ein Redner", "Der Student", "Der Bürger", and "alle Kollegen" are masculine [M];
    \\- Phrases like "Eine Rednerin", "Die Studentin", "Die Bürgerinnen", and "alle Kolleginnen" are feminine [F];
    \\- Phrases like "Eine referierende Person", "Die Studierenden", "Die Bürgerschaft", and "alle Kollegiumsmitgliedern" do not express social gender, therefore they must be considered neutral [N].\\
    
    \\3. Assign a sentence-level label:
    \\- If all references to human beings are gender-neutral, label the sentence as "NEUTRAL".
    \\- If one or more phrases convey a specific masculine or feminine gender, label the sentence as "GENDERED".
   \\
   \hline
   
    \end{tabularx}
    \caption{System message for prompt \textsc{Mono-P+L} (German).}
    \label{tab:prompt-Mono-P+L}
\end{table*}

\begin{table*}
\small
    \centering
    \begin{tabularx}{\textwidth}{|X|}
    \hline
     You are a language expert specializing in evaluating gender-neutral translation from English into Spanish. Your task is to assess each provided source-target sentence pair and determine whether the sentence was translated in a correctly gendered, wrongly gendered, or neutral way. \\

    \\Guidelines:\\
    \\1. Identify relevant phrases: carefully read the Spanish sentence and identify all phrases that refer to human beings or groups of human beings, including:\\
    \\- Noun phrases (e.g., "una excelente oradora", "la ciudadanía"),
    \\- Verb phrases (e.g., "es muy feliz", "lamentablemente tuve que hacerlo"),
    \\- Adjective phrases (e.g., "felizmente casado", "muy competente").\\
    
    \\2. Evaluate gender information: consider only the social gender conveyed by the phrases, not grammatical gender. For example:
    \\- Phrases like "un orador", "es muy contento", "todos los colegas", and "los ciudadanos" are masculine;
    \\- Phrases like "una oradora", "es muy contenta", "todas las colegas", and "las ciudadanas" are feminine;
    \\- Phrases like "una persona que habla en público", "es muy feliz", "todas las personas con las que trabajo", and "la ciudadanía" do not express social gender, therefore they must be considered neutral.\\
    
    \\3. Assess gender correctness: for each extracted phrase, assess the correctness of the social gender expressed in the Spanish phrase based on the information available in the source English sentence. Consider that:\\
    \\- Masculine phrases must correspond to masculine gender cues in English (e.g., he, him, Mr, man) to be considered correct.
    \\- Feminine phrases must correspond to feminine gender cues in English (e.g., she, her, Ms, woman) to be considered correct.
    \\- Neutral phrases do not need to be matched with gender cues in the source to be correct. Note that proper names do not count as valid gender cues, ignore them.\\
    
    \\4. Assign a label to the translation:
    \\- If there are masculine or feminine phrases in the Spanish text and the source contains matching gender cues, label the sentence as "CORRECTLY GENDERED".
    \\- If there are masculine or feminine phrases in the Spanish text and the source does not contain matching gender cues, label the sentence as "WRONGLY GENDERED".
    \\- If there are only neutral phrases in the Spanish text, label the sentence as "NEUTRAL".
   \\
   \hline
   
    \end{tabularx}
    \caption{System message for prompt \textsc{Mono-L} (Spanish).}
    \label{tab:prompt-Cross-L}
\end{table*}

\begin{table*}
\small
    \centering
    \begin{tabularx}{\textwidth}{|X|}
    \hline
     You are an expert language annotator and evaluator of gender-neutral translation for English-Italian. Your task is to extract target Italian phrases that refer to human beings, determine whether each phrase is masculine, feminine, or neutral, and assess if the gender expressed in each phrase is correct with respect to the source. Based on the phrases, determine whether the sentence was translated in a correctly gendered, wrongly gendered, or neutral way.\\

    \\Guidelines:\\
    \\1. Identify relevant phrases: carefully read the Italian sentence and extract all phrases that refer to human beings or groups of human beings, including:\\
    \\- Noun phrases (e.g., "un’ottima oratrice", "la cittadinanza"),
    \\- Verb phrases (e.g., "è molto felice", "ho purtroppo dovuto"),
    \\- Adjective phrases (e.g., "felicemente sposato", "molto competente").\\
    
    \\2. Evaluate gender information: consider only the social gender conveyed by the phrases, not grammatical gender, and assign a label to each phrase [M/F/N]. For example:
    \\- Phrases like "un oratore", "è molto contento", "tutti i colleghi", and "i cittadini" are masculine [M];
    \\- Phrases like "un’oratrice", "è molto contenta", "tutte le colleghe", and "le cittadine" are feminine [F];
    \\- Phrases like "una persona che parla in pubblico", "è molto felice", "tutte le persone con cui lavoro", and "la cittadinanza" do not express social gender, therefore they must be considered neutral [N].\\
    
    \\3. Assess gender correctness: for each extracted phrase, assess the correctness of the social gender expressed in the Italian phrase based on the information available in the source English sentence [correct/wrong]. Consider that:
    \\- If a phrase is masculine, the English source must contain masculine gender cues (e.g., he, him, Mr, man) for it to be correct.
    \\- If a phrase is feminine, the English source must contain feminine gender cues (e.g., she, her, Ms, woman) for it to be correct.
    \\- If a phrase is neutral, it is always correct, regardless of gender cues in the source. Note that proper names do not count as valid gender cues, ignore them.
    
    \\4. Assign a sentence-level label to the translation:
    \\- If there are masculine or feminine phrases in the Italian text and the source contains matching gender cues, label the sentence as "CORRECTLY GENDERED".
    \\- If there are masculine or feminine phrases in the Italian text and the source does not contain matching gender cues, label the sentence as "WRONGLY GENDERED".
    \\- If there are only neutral phrases in the Italian text, label the sentence as "NEUTRAL".

   \\
   \hline
   
    \end{tabularx}
    \caption{System message for prompt \textsc{Mono-P+L} (Italian).}
    \label{tab:prompt-Cross-P+L}
\end{table*}




\section{Complete results}
\label{app:results}

This section contains the detailed results of the evaluation experiments introduced in Section \ref{sec:experimental-settings}. Tables \ref{tab:mono-results-it}, \ref{tab:mono-results-de}, and \ref{tab:mono-results-es} report the 
\ap{accuracy of all models}
in \textit{target-only} evaluation of Italian, German, and Spanish mGeNTE references respectively, whereas 
Table \ref{tab:eacl-results} reports 
\ap{their precision, recall, and F1 scores in}
the evaluation of Italian automatic GNTs.
Tables \ref{tab:cross-results-it}, \ref{tab:cross-results-de}, and \ref{tab:cross-results-es} report 
\ap{models' accuracy}
in GNT evaluation.

While in Figures \ref{fig:mono-results} and \ref{fig:cross-results} we aggregated the results each model/prompt combination by Set (G or N), here we report the results of the experiments on mGeNTE references on each set/reference split too, as well as their average. This allows for the analysis of models' accuracy in each configuration on the different references.

By comparing models' performance on \texttt{REF-G} and \texttt{REF-N}, we notice a general gap between performance on the first versus the latter, especially for the label-only prompts.
The instances where the simpler 
prompts
result in higher scores on \texttt{REF-G} are due to models' inability to recognize neutral phrases, which causes them to default to the 
\fcolorbox{Lavender!20}{Lavender!20}{\texttt{\color{violet}{GENDERED}}} label(s). This is reflected in their low scores on \texttt{REF-N}, which ultimately sinks the overall performance of those prompts shown in the charts. When guided towards the generation of richer annotations before providing sentence-level assessments, the performance of all models on \texttt{REF-N} improves significantly, with a small impact on \texttt{REF-G} performance. This improvement is the main driver of the higher overall accuracy.

This behavior reflects the one we noticed and discussed in \ref{sec:results-cross}, and confirm models' tendency to generate the 
\fcolorbox{Lavender!20}{Lavender!20}{\texttt{\color{violet}{GENDERED}}} 
label(s) rather than 
\fcolorbox{Turquoise!20}{Turquoise!20}{\texttt{\color{Blue}{NEUTRAL}}} 
on mGeNTE references as well.

\setlength{\tabcolsep}{3.5pt}
\begin{table*}
\small
    \centering
    \begin{tabular}{lc|cccc|cccc|cccc}
        \toprule
        \multicolumn{2}{c|}{\textbf{en-it}} & \multicolumn{4}{c|}{\textbf{REF-G}} & \multicolumn{4}{c|}{\textbf{REF-N}}  & \multicolumn{4}{c}{\textbf{OVERALL}}\\
        \midrule
        \textbf{SYSTEM} & \textbf{SPLIT} & \ding{109} & \ding{108} &   $\lozenge$ & $\blacklozenge$ &  \ding{109} & \ding{108} &   $\lozenge$ & $\blacklozenge$ & \ding{109} & \ding{108} &   $\lozenge$ & $\blacklozenge$ \\
        \midrule
        \multirow{3}{*}{GPT-4o} 
        & Set-G  & \underline{96.38}  & \underline{99.06}  & \cellcolor{yellow!40} \underline{99.73} & \underline{99.33} & 52.95 & 61.66 & 4.02 & \cellcolor{yellow!40} \underline{\textbf{83.11}} & 74.67 & \underline{80.36} & 51.88 & \cellcolor{yellow!40} \underline{\textbf{91.22}}  \\
        & Set-N  & 62.33  & \underline{89.54}  & \underline{88.34} & \cellcolor{yellow!40} \underline{90.21} & 89.28 & 86.73 & 18.36 & \cellcolor{yellow!40} \underline{\textbf{90.21}} & 75.81 & \underline{88.14} & 53.35 & \cellcolor{yellow!40} \underline{\textbf{90.21}} \\
        & Overall & 79.36  & \underline{94.30}  & \underline{94.03} & \cellcolor{yellow!40} \underline{94.77} & 71.12 & 74.20 & 11.19 & \cellcolor{yellow!40} \underline{\textbf{86.66}} & 75.24 & \underline{84.25} & 52.61 & \cellcolor{yellow!40} \underline{\textbf{90.72}} \\
        \midrule
        \multirow{3}{*}{Qwen 32B} 
        & Set-G  & \underline{99.06}  & \underline{98.93}  & \underline{98.93} & \cellcolor{yellow!40} \underline{99.46} & 21.18 & 33.78 & 1.74 & \cellcolor{yellow!40} 34.45 & 60.12 & 66.36 & 50.34 & \cellcolor{yellow!40} 66.96 \\
        & Set-N  & \underline{82.71}  & \underline{91.82}  & \underline{93.16} & \cellcolor{yellow!40} \underline{93.70} & 48.12 & 55.90 & 8.98 & \cellcolor{yellow!40} 60.19 & 65.42 & 73.86 & 51.07 & \cellcolor{yellow!40} 76.95 \\
        & Overall & \underline{90.89}  & \underline{95.38} & \underline{96.05} & \cellcolor{yellow!40} \underline{96.58} & 34.65 & 44.84 & 5.36 & \cellcolor{yellow!40} 47.32 & 62.77 & 70.11 & 50.71 & \cellcolor{yellow!40} 71.95 \\
        \midrule
        \multirow{3}{*}{Qwen 72B} 
        & Set-G  & \cellcolor{yellow!40} \textbf{\underline{100.00}}  & \underline{98.79} & \underline{98.66} & \underline{98.66} & 1.12 & 43.70 & 2.95 & \cellcolor{yellow!40} 61.93 & 50.61 & 71.25 & 50.81 & \cellcolor{yellow!40} \underline{80.30}\\
        & Set-N  & \cellcolor{yellow!40} \textbf{\underline{96.65}}  & \underline{83.24} & \underline{87.94} & \underline{80.56} & 5.76 & 76.54 & 11.13 & \cellcolor{yellow!40} 80.29 & 51.21 & 79.89 & 49.54 & \cellcolor{yellow!40} 80.43  \\
        & Overall & \cellcolor{yellow!40} \textbf{\underline{98.33}}  & \underline{91.02} & \underline{93.30} & \underline{89.61} & 3.49 & 60.12 & 7.04 & \cellcolor{yellow!40} 71.11 & 50.91 & 75.57 & 50.17 & \cellcolor{yellow!40} 80.46 \\
        \midrule
        \multirow{3}{*}{Mistral Small} 
        & Set-G  & \underline{98.12}  & \underline{99.33} & \underline{98.93} & \cellcolor{yellow!40} \underline{99.73} & 10.05 & 23.06 & 0.67 & \cellcolor{yellow!40} 3.24 & 54.09 & 61.20 & 49.80 & \cellcolor{yellow!40} 66.49 \\
        & Set-N  & \underline{77.88}  & \underline{90.48} & \cellcolor{yellow!40} \underline{95.98} & \underline{93.97} & 41.55 & \cellcolor{yellow!40} 54.69 & 5.09 & 53.75 & 59.72 & 72.59 & 50.54 & \cellcolor{yellow!40} 73.86 \\
        & Overall & \underline{88.00}  & \underline{94.91} & \cellcolor{yellow!40} \underline{97.45}& \underline{96.85} & 25.80 & 38.88 & 2.88 & \cellcolor{yellow!40} 43.50 & 56.90 & 66.89 & 50.17 & \cellcolor{yellow!40} 70.17 \\
        \midrule
        \multirow{3}{*}{DS Qwen 32B} 
        & Set-G  & \underline{98.12}  & \cellcolor{yellow!40} \underline{99.20}  & \underline{94.91} & \underline{99.06} & 8.58 & 23.86 & 6.84 & \cellcolor{yellow!40} 26.68 & 53.82 & 61.53 & 50.88 & \cellcolor{yellow!40} 62.87 \\
        & Set-N  & \underline{77.88}  & \underline{91.82}  & \underline{86.60} & \cellcolor{yellow!40} \underline{93.57} & 31.10 & \cellcolor{yellow!40} 47.72 & 16.16 & 45.04 & 58.98 & \cellcolor{yellow!40} 69.77 & 51.88 & 69.31  \\
        & Overall & \underline{88.00}  & \underline{95.51}  & \underline{90.75} & \cellcolor{yellow!40} \underline{96.32} & 19.84 & 35.79 & 12.00 & \cellcolor{yellow!40} 35.86 & 56.40 & 65.65 & 51.38 & \cellcolor{yellow!40} 66.09 \\
        \midrule
        \midrule
        \multirow{3}{*}{Classifier}
        & Set-G  & \multicolumn{4}{c|}{92.76} & \multicolumn{4}{c|}{66.49} & \multicolumn{4}{c}{79.63} \\
        & Set-N & \multicolumn{4}{c|}{76.81} & \multicolumn{4}{c|}{89.41} & \multicolumn{4}{c}{83.11} \\
        & Overall & \multicolumn{4}{c|}{84.79} & \multicolumn{4}{c|}{77.95} & \multicolumn{4}{c}{81.37} \\

        \bottomrule

    \end{tabular}
    \caption{
    \ap{Accuracy of all models in }
    \textit{target-only} English~→~Italian GNT evaluation on mGeNTE references, including those of the gender-neutrality classifier \cite{savoldi-etal-2024-prompt}, which acts as a baseline for these experiments. Instances where models outperform the classifier in a specific data split are underlined. The best-performing settings for each data split are in bold. The best performing strategy per model and data split is highlighted.}
    \label{tab:mono-results-it}
\end{table*}

\begin{table*}
\small
    \centering
    \begin{tabular}{lc|cccc|cccc|cccc}
        \toprule
        \multicolumn{2}{c|}{\textbf{en-de}} & \multicolumn{4}{c|}{\textbf{REF-G}} & \multicolumn{4}{c|}{\textbf{REF-N}}  & \multicolumn{4}{c}{\textbf{OVERALL}}\\
        \midrule
        \textbf{SYSTEM} & \textbf{SPLIT} & \ding{109} & \ding{108} &   $\lozenge$ & $\blacklozenge$ &  \ding{109} & \ding{108} &   $\lozenge$ & $\blacklozenge$ & \ding{109} & \ding{108} &   $\lozenge$ & $\blacklozenge$ \\
        \midrule
        \multirow{3}{*}{GPT-4o} 
        & Set-G  & 99.06  & \cellcolor{yellow!40} 99.73  & 96.38 & 99.60 & 80.56 & 78.82 & 65.28 & \cellcolor{yellow!40} 81.37 & 89.81 & 89.28 & 80.83 & \cellcolor{yellow!40} \textbf{90.49}  \\
        & Set-N  & 81.10  & \cellcolor{yellow!40} 88.47  & 66.89 & \cellcolor{yellow!40} 88.47 & 90.48 & 95.84 & 95.58 & \cellcolor{yellow!40} 97.59 & 85.79 & 92.16 & 81.24 & \cellcolor{yellow!40} \textbf{93.03}  \\
        & Overall & 90.08  & \cellcolor{yellow!40} 94.10  & 81.64 & 94.03 & 85.52 & 87.33 & 80.43 & \cellcolor{yellow!40} 89.48 & 87.80 & 90.72 & 81.03 & \cellcolor{yellow!40} \textbf{91.76} \\
        \midrule
        \multirow{3}{*}{Qwen 32B} 
        & Set-G  & \cellcolor{yellow!40} 99.73 & 99.60 & 95.04 & 99.73& 7.64 & 49.87 & 38.07 & \cellcolor{yellow!40} 54.69 & 53.69 & 74.74 & 66.56 & \cellcolor{yellow!40}77.21  \\
        & Set-N  & \cellcolor{yellow!40} 96.65  & 90.08 & 66.22 & 84.99 & 13.67 & 59.52 & 59.52 & \cellcolor{yellow!40} 76.01 & 55.16 & 74.80 & 62.87 & \cellcolor{yellow!40} 80.50  \\
        & Overall & \cellcolor{yellow!40} 98.19 & 94.84 & 80.63 & 92.36 & 10.66	& 54.69 & 48.79 & \cellcolor{yellow!40} 65.35 & 54.42 & 74.77 & 64.71 & \cellcolor{yellow!40} 78.86  \\
        \midrule
        \multirow{3}{*}{Qwen 72B} 
        & Set-G  & \cellcolor{yellow!40} 100.00 & 99.60 & 98.66 & 99.46 & 1.21 & 62.06 & 21.31 & \cellcolor{yellow!40} 81.90 & 50.61 & 80.83 & 59.99 & \cellcolor{yellow!40} 90.68  \\
        & Set-N  & \cellcolor{yellow!40} 99.73 & 82.71 & 63.00 & 75.60 & 1.61 & 86.33 & 57.77 & \cellcolor{yellow!40} 93.03 & 50.67 & 84.52 & 60.39 & \cellcolor{yellow!40} 84.32  \\
        & Overall & \cellcolor{yellow!40} 99.87 & 91.15 & 80.83 & 87.53 & 1.41 & 74.20 & 39.54 & \cellcolor{yellow!40} 87.47 & 50.64 & 82.68 & 60.19 & \cellcolor{yellow!40} 87.50  \\
        \midrule
        \multirow{3}{*}{Mistral Small} 
        & Set-G  & 98.12 & \cellcolor{yellow!40} 99.60 & 96.38 & 99.46 & 15.55 & \cellcolor{yellow!40} 57.85 & 33.51 & 52.82 & 54.16 & 69.57 & 60.12 & \cellcolor{yellow!40} 69.71  \\
        & Set-N  & 79.09 & 88.47 & 56.30 & \cellcolor{yellow!40} 94.64 & 35.92 & \cellcolor{yellow!40} 81.23 & 64.34 & 74.66 & 57.44 & \cellcolor{yellow!40} 78.29 & 63.14 & 74.94  \\
        & Overall & 88.61 & 94.03 & 76.34 & \cellcolor{yellow!40} 97.05 & 25.74 & \cellcolor{yellow!40} 69.55 & 48.93 & 63.74 & 55.80 & \cellcolor{yellow!40} 73.93 & 61.63 & 72.32  \\
        \midrule
        \multirow{3}{*}{DS Qwen 32B} 
        & Set-G  & \cellcolor{yellow!40} 99.73 & 99.33 & 91.69 & 99.60 & 1.47 & \cellcolor{yellow!40} 52.89 & 19.57 & 44.64 & 50.60 & \cellcolor{yellow!40} 76.11 & 55.63 & 72.12  \\
        & Set-N  & \cellcolor{yellow!40} 95.98 & 92.63 & 70.24 & 92.63 & 9.52 & 53.08 & \cellcolor{yellow!40} 54.56 & 46.18 & 52.75	 & \cellcolor{yellow!40} 72.86 & 62.40 & 69.41  \\
        & Overall & \cellcolor{yellow!40} 97.86 & 95.98 & 80.97 & 96.11 & 5.50 & \cellcolor{yellow!40} 52.98 & 37.06 & 45.41 & 51.68 & \cellcolor{yellow!40} 74.48 & 59.0 & 70.76  \\
        \bottomrule

    \end{tabular}
    \caption{
    \ap{Accuracy of all models in}
    \textit{target-only} English~→~German GNT evaluation on mGeNTE references. The best-performing settings for each data split are in bold. The best performing strategy per model and data split is highlighted.}
    \label{tab:mono-results-de}
\end{table*}

\begin{table*}
\small
    \centering
    \begin{tabular}{lc|cccc|cccc|cccc}
        \toprule
        \multicolumn{2}{c|}{\textbf{en-es}} & \multicolumn{4}{c|}{\textbf{REF-G}} & \multicolumn{4}{c|}{\textbf{REF-N}}  & \multicolumn{4}{c}{\textbf{OVERALL}}\\
        \midrule
        \textbf{SYSTEM} & \textbf{SPLIT} & \ding{109} & \ding{108} &   $\lozenge$ & $\blacklozenge$ &  \ding{109} & \ding{108} &   $\lozenge$ & $\blacklozenge$ & \ding{109} & \ding{108} &   $\lozenge$ & $\blacklozenge$ \\
        \midrule
        \multirow{3}{*}{GPT-4o} 
        & Set-G  & 98.39 & 99.73 & 95.71 & \cellcolor{yellow!40} 99.87 & 62.33 & 74.13 & 62.33 & \cellcolor{yellow!40} 90.75 & 80.36 & 86.93 & 79.02 & \cellcolor{yellow!40} \textbf{95.31}  \\
        & Set-N  & 70.11 & 91.42 & 46.38 & \cellcolor{yellow!40} 95.58 & 86.86 & 91.02 & \cellcolor{yellow!40} 96.11 & 94.10 & 78.49 & 91.22 & 71.25 & \cellcolor{yellow!40} \textbf{94.84}  \\
        & Overall & 84.25 & 95.58 & 71.05 & \cellcolor{yellow!40} 97.72 & 74.60 & 82.57 & 79.22 & \cellcolor{yellow!40} 92.43 & 79.43 & 89.08 & 75.14 & \cellcolor{yellow!40} \textbf{95.08 } \\
        \midrule
        \multirow{3}{*}{Qwen 32B} 
        & Set-G  & 98.93 & \cellcolor{yellow!40} 99.60 & 94.10 & \cellcolor{yellow!40} 99.60 & 21.72 & 33.65 & \cellcolor{yellow!40} 58.71 & 58.45 & 60.33 & 66.63 & 76.41 & \cellcolor{yellow!40} 79.03  \\
        & Set-N  & 83.65 & 96.25 & 60.05 & \cellcolor{yellow!40} 96.78 & 51.07 & 52.85 & \cellcolor{yellow!40} 78.95 & 61.39 & 67.36 & 74.55 & 69.50 & \cellcolor{yellow!40} 79.09  \\
        & Overall & 91.29 & 97.29 & 77.08 & \cellcolor{yellow!40} 98.19 & 36.39 & 43.23 & \cellcolor{yellow!40} 68.83 & 59.62 & 63.84 & 70.59 & 72.95 & \cellcolor{yellow!40} 78.91  \\
        \midrule
        \multirow{3}{*}{Qwen 72B} 
        & Set-G  & \cellcolor{yellow!40} \textbf{100.00} & 99.73 & 99.06 & 99.33 & 1.07 & 48.66 & 12.06 & \cellcolor{yellow!40} 65.01 & 50.54 & 74.20 & 55.56 & \cellcolor{yellow!40} 82.17  \\
        & Set-N  & \cellcolor{yellow!40} \textbf{98.93} & 87.27 & 57.64 & 85.52 & 3.62 & \cellcolor{yellow!40} 82.98 & 57.37 & 82.04 & 51.28 & \cellcolor{yellow!40} 85.13 & 57.51 & 83.78  \\
        & Overall & \cellcolor{yellow!40} \textbf{99.46} & 93.50 & 78.35 &92.43 & 2.35 & 65.28 & 34.72 & \cellcolor{yellow!40} 73.53 & 50.91 & 79.39 & 56.54 & \cellcolor{yellow!40} 82.98  \\
        \midrule
        \multirow{3}{*}{Mistral Small} 
        & Set-G  & 98.12 & \cellcolor{yellow!40} 99.60 & 96.38 & 99.46 & 10.19 & 39.54 & 23.86 & \cellcolor{yellow!40} 39.95 & 54.16 & 69.57 & 60.12 & \cellcolor{yellow!40} 69.71  \\
        & Set-N  & 79.09 & 88.47 & 56.30 & \cellcolor{yellow!40} 94.64 & 35.79 & 68.10 & \cellcolor{yellow!40} 69.97 & 55.23 & 57.44 & \cellcolor{yellow!40} 78.29 & 63.14 & 74.94  \\
        & Overall & 88.61 & 94.03 & 76.34 & \cellcolor{yellow!40} 97.05 & 22.99 & 53.82 & 46.92 & \cellcolor{yellow!40} 47.59 & 55.80 & \cellcolor{yellow!40} 73.93 & 61.63 & 72.32  \\
        \midrule
        \multirow{3}{*}{DS Qwen 32B} 
        & Set-G  & 98.53 & \cellcolor{yellow!40} 99.46 & 83.11 & 99.33 & 5.76 & 40.35 & 32.04 & \cellcolor{yellow!40} 51.34 & 52.15 & 69.91 & 57.58 & \cellcolor{yellow!40} 75.34  \\
        & Set-N  & 90.75 & 93.83 & 61.80 & \cellcolor{yellow!40} 96.51 & 22.79 & \cellcolor{yellow!40} 52.95 & 50.94 & 50.00 & 56.77 & \cellcolor{yellow!40} 73.39 & 56.37 & 73.26  \\
        & Overall & 94.64 & 96.65 & 72.45 & \cellcolor{yellow!40}97.92 & 14.28 & 46.65 & 41.49 & \cellcolor{yellow!40} 50.67 & 54.46 & 71.65 & 56.97 & \cellcolor{yellow!40} 74.30  \\
        \bottomrule

    \end{tabular}
    \caption{
    \ap{Accuracy of all models in}
    \textit{target-only} English~→~Spanish GNT evaluation on mGeNTE references. The best-performing settings for each data split are in bold. The best performing strategy per model and data split is highlighted.}
    \label{tab:mono-results-es}
\end{table*}

\begin{table*}
\small
    \centering
    \begin{tabular}{l|cccc|cccc|cccc}
        \toprule
        \textbf{Italian} & \multicolumn{4}{c|}{\textbf{Precision}} & \multicolumn{4}{c|}{\textbf{Recall}}  & \multicolumn{4}{c}{\textbf{F1}}\\
        \midrule
        \textbf{SYSTEM} &  \ding{109} & \ding{108} &   $\lozenge$ & $\blacklozenge$ &  \ding{109} & \ding{108} &   $\lozenge$ & $\blacklozenge$ & \ding{109} & \ding{108} &   $\lozenge$ & $\blacklozenge$ \\
        
        \midrule
        GPT-4o & 72.58 & \cellcolor{yellow!40} \underline{81.17} & 46.61 & \underline{80.07} & \cellcolor{yellow!40} \underline{\textbf{96.22}} & 93.78 & 7.43 & \underline{95.00} & 82.74 & \cellcolor{yellow!40} \underline{\textbf{87.02}} & 12.82 & \underline{86.90}  \\
        Qwen 32B & 75.05 & 73.19 & 75.42 & \cellcolor{yellow!40} \underline{80.03} & 51.62 & 61.62 & \cellcolor{yellow!40} 78.38 & 74.05 & 61.17 & 67.21 & 76.87 & \cellcolor{yellow!40} 77.29  \\
        Qwen 72B & \cellcolor{yellow!40} \underline{\textbf{93.33}} & 77.95 & 72.32 & 75.95 & 7.57 & 88.38 & 85.81 & \cellcolor{yellow!40} 89.59 & 14.00 & \cellcolor{yellow!40} 82.84 & 78.49 & 82.21  \\
        Mistral Small & \cellcolor{yellow!40} \underline{81.22} & 78.94 & 75.70 & 78.85 & 52.03 & 70.41 & \cellcolor{yellow!40} 84.19 & 72.03 & 63.43 & 74.43 & \cellcolor{yellow!40} 79.72 & 75.28  \\
        DS Qwen 32B & \cellcolor{yellow!40} 73.47 & 72.90 & 71.60 & 73.11 & 29.19 & 51.62 & \cellcolor{yellow!40} 62.70 & 48.11 & 41.78 & 60.44 & \cellcolor{yellow!40} 66.86 & 58.03  \\
        \midrule
        \midrule
        Classifier & \multicolumn{4}{c|}{79.36} & \multicolumn{4}{c|}{94.05} & \multicolumn{4}{c}{86.09}  \\
        \bottomrule

    \end{tabular}
    \caption{
    \ap{Precision, recall, and F1 scores of all models in} 
    \textit{target-only} English~→~Italian GNT evaluation on automatic GNTs, including those of the gender-neutrality classifier \cite{savoldi-etal-2024-prompt}, which acts as a baseline for these experiments. Instances where models outperform the classifier are underlined. The best-performing settings are in bold. The best performing strategy per model is highlighted.}
    \label{tab:eacl-results}
\end{table*}

\begin{table*}
\small
    \centering
    \begin{tabular}{lc|cc|cc|cc}
        \toprule
        \multicolumn{2}{c|}{\textbf{en-it}} & \multicolumn{2}{c|}{\textbf{REF-G}} & \multicolumn{2}{c|}{\textbf{REF-N}}  & \multicolumn{2}{c}{\textbf{OVERALL}}\\
        \midrule
        \textbf{SYSTEM} & \textbf{SPLIT} &   $\lozenge$ & $\blacklozenge$ &   $\lozenge$ & $\blacklozenge$ &   $\lozenge$ & $\blacklozenge$ \\
        \midrule
        \multirow{3}{*}{GPT-4o} 
        & Set-G  & 73.99 & \cellcolor{yellow!40} 84.32  & 4.02 & \cellcolor{yellow!40} \textbf{83.11} & 39.01 & \cellcolor{yellow!40} \textbf{83.72} \\
        & Set-N  & 34.18 & \cellcolor{yellow!40} 51.88  & 18.36 & \cellcolor{yellow!40} \textbf{90.21} & 26.27 & \cellcolor{yellow!40} 71.05 \\
        & Overall & 54.09 & \cellcolor{yellow!40} 68.10  & 11.19 & \cellcolor{yellow!40} \textbf{86.66} & 32.64 & \cellcolor{yellow!40} \textbf{77.38} \\
        
        \midrule
        \multirow{3}{*}{Qwen 32B} 
        & Set-G  & \cellcolor{yellow!40} \textbf{94.50} & 81.23  & 1.74 & \cellcolor{yellow!40} 34.45 & 48.12 & \cellcolor{yellow!40} 57.84 \\
        & Set-N  & 14.21 & \cellcolor{yellow!40} 90.32  & 8.98 & \cellcolor{yellow!40} 60.19 & 11.60 & \cellcolor{yellow!40} \textbf{75.26} \\
        & Overall & 54.36 & \cellcolor{yellow!40} \textbf{85.78}  & 5.36 & \cellcolor{yellow!40} 47.32 & 29.86 & \cellcolor{yellow!40} 66.55 \\
        
        \midrule
        \multirow{3}{*}{Qwen 72B} 
        & Set-G  & 88.07 & \cellcolor{yellow!40} 92.90  & 2.95 & \cellcolor{yellow!40} 61.93 & 45.51 & \cellcolor{yellow!40} 77.42 \\
        & Set-N  & 15.82 & \cellcolor{yellow!40} 65.82  & 11.13 & \cellcolor{yellow!40} 80.29 & 13.48 & \cellcolor{yellow!40} 73.06 \\
        & Overall & 51.95 & \cellcolor{yellow!40} 79.36  & 7.04	& \cellcolor{yellow!40} 71.11 & 29.49 & \cellcolor{yellow!40} 75.24 \\
        
        \midrule
        \multirow{3}{*}{Mistral Small} 
        & Set-G  & \cellcolor{yellow!40} 74.40 & 68.77  & 0.67 & \cellcolor{yellow!40} 33.42 & 37.54 & \cellcolor{yellow!40} 51.10 \\
        & Set-N  & 43.30 & \cellcolor{yellow!40} \textbf{92.76}  & 5.09	& \cellcolor{yellow!40} 53.75 & 24.20 & \cellcolor{yellow!40} 73.26 \\
        & Overall & 58.85 &\cellcolor{yellow!40}  80.77  & 2.88 & \cellcolor{yellow!40} 43.59 & 30.87 & \cellcolor{yellow!40} 60.74 \\
        
        \midrule
        \multirow{3}{*}{DS Qwen 32B} 
        & Set-G  & \cellcolor{yellow!40} 85.25 & 84.85  & 6.84	 & \cellcolor{yellow!40} 6.68 & 46.05 & \cellcolor{yellow!40} 55.77 \\
        & Set-N  & 3.62	& \cellcolor{yellow!40} 85.79  & 17.16 & \cellcolor{yellow!40} 45.04 & 10.39 & \cellcolor{yellow!40} 65.42 \\
        & Overall & 44.44 & \cellcolor{yellow!40} 85.32  & 12.00 & \cellcolor{yellow!40} 35.86 & 28.22 & \cellcolor{yellow!40} 60.59 \\
        \bottomrule

    \end{tabular}
    \caption{
    \ap{Accuracy of all models in}
    \textit{source-target}  English~→~Italian GNT evaluation on mGeNTE source-reference pairs. The best-performing settings for each data split are in bold. The best performing strategy per model and data split is highlighted.}
    \label{tab:cross-results-it}
\end{table*}

\begin{table*}
\small
    \centering
    \begin{tabular}{lc|cc|cc|cc}
        \toprule
        \multicolumn{2}{c|}{\textbf{en-de}} & \multicolumn{2}{c|}{\textbf{REF-G}} & \multicolumn{2}{c|}{\textbf{REF-N}}  & \multicolumn{2}{c}{\textbf{OVERALL}}\\
        \midrule
        \textbf{SYSTEM} & \textbf{SPLIT} &   $\lozenge$ & $\blacklozenge$ &   $\lozenge$ & $\blacklozenge$ &   $\lozenge$ & $\blacklozenge$ \\
        \midrule
        \multirow{3}{*}{GPT-4o} 
        & Set-G  & 59.92 & \cellcolor{yellow!40} 76.54 & 65.28 & \cellcolor{yellow!40} 81.37 & 62.60 & \cellcolor{yellow!40} 78.96 \\
        & Set-N  & 64.21 & \cellcolor{yellow!40} 81.37 & 95.58 & \cellcolor{yellow!40} \textbf{97.59} & 79.90 & \cellcolor{yellow!40} \textbf{89.48} \\
        & Overall & 62.07 & \cellcolor{yellow!40} 78.96 & 80.43 & \cellcolor{yellow!40} \textbf{89.48} & 71.25 & \cellcolor{yellow!40} \textbf{84.22} \\
        
        \midrule
        \multirow{3}{*}{Qwen 32B} 
        & Set-G  & \cellcolor{yellow!40} 87.40 & 85.12 & 38.07 & \cellcolor{yellow!40} 54.69 & 62.74 & \cellcolor{yellow!40} 69.91 \\
        & Set-N  & 24.53 & \cellcolor{yellow!40} 79.89 & 59.52 & \cellcolor{yellow!40} 76.01 & 42.03 & \cellcolor{yellow!40} 77.95 \\
        & Overall & 55.97 & \cellcolor{yellow!40} 82.51 & 48.80 & \cellcolor{yellow!40} 65.35 & 52.38 & \cellcolor{yellow!40} 73.93 \\
        
        \midrule
        \multirow{3}{*}{Qwen 72B} 
        & Set-G  & 66.62 & \cellcolor{yellow!40} \textbf{91.42} & 21.31 & \cellcolor{yellow!40} \textbf{81.90} & 43.97 & \cellcolor{yellow!40} \textbf{86.66} \\
        & Set-N  & 47.99 & \cellcolor{yellow!40} 63.54 & 57.77 & \cellcolor{yellow!40} 93.03 & 52.88 & \cellcolor{yellow!40} 78.29 \\
        & Overall & 57.31 & \cellcolor{yellow!40} 77.48 & 39.54 & \cellcolor{yellow!40} 87.47 & 48.42 & \cellcolor{yellow!40} 82.48 \\
        
        \midrule
        \multirow{3}{*}{Mistral Small} 
        & Set-G  & \cellcolor{yellow!40} 77.35 & 69.80 & 33.51 & \cellcolor{yellow!40} 52.82 & 55.43 & \cellcolor{yellow!40} 61.31 \\
        & Set-N  & 53.62 & \cellcolor{yellow!40} \textbf{83.78} & 64.34 & \cellcolor{yellow!40} 74.66 & 58.98 & \cellcolor{yellow!40} 79.22 \\
        & Overall & 65.49 & \cellcolor{yellow!40} 76.79 & 48.93 & \cellcolor{yellow!40} 63.74 & 57.21 & \cellcolor{yellow!40} 70.27 \\

        \midrule
        \multirow{3}{*}{DS Qwen 32B} 
        & Set-G  & 32.57 & \cellcolor{yellow!40} 87.53  & 19.57 & \cellcolor{yellow!40} 44.64 & 26.07 & \cellcolor{yellow!40} 66.09 \\
        & Set-N  & 41.29 & \cellcolor{yellow!40} 82.17  & \cellcolor{yellow!40} 54.56 & 46.18 & 47.93 & \cellcolor{yellow!40} 64.18 \\
        & Overall & 36.93 & \cellcolor{yellow!40} \textbf{84.85}  & 37.07 & \cellcolor{yellow!40} 45.41 & 37.00 & \cellcolor{yellow!40} 65.13 \\

        \bottomrule

    \end{tabular}
    \caption{
    \ap{Accuracy of all models in}
    \textit{source-target} English~→~German GNT evaluation on mGeNTE source-reference pairs. The best-performing settings for each data split are in bold. The best performing strategy per model and data split is highlighted.}
    \label{tab:cross-results-de}
\end{table*}

\begin{table*}
\small
    \centering
    \begin{tabular}{lc|cc|cc|cc}
        \toprule
        \multicolumn{2}{c|}{\textbf{en-es}} & \multicolumn{2}{c|}{\textbf{REF-G}} & \multicolumn{2}{c|}{\textbf{REF-N}}  & \multicolumn{2}{c}{\textbf{OVERALL}}\\
        \midrule
        \textbf{SYSTEM} & \textbf{SPLIT} &   $\lozenge$ & $\blacklozenge$ &   $\lozenge$ & $\blacklozenge$ &   $\lozenge$ & $\blacklozenge$ \\
        \midrule
        \multirow{3}{*}{GPT-4o} 
        & Set-G  & 67.43  & \cellcolor{yellow!40}79.09  & 62.33  & \cellcolor{yellow!40}\textbf{90.75} & 64.88 & \cellcolor{yellow!40}\textbf{84.92} \\
        & Set-N  & 42.76  & \cellcolor{yellow!40}58.98  & \cellcolor{yellow!40}\textbf{96.11}  & 94.10 & 69.44 & \cellcolor{yellow!40}76.54 \\
        & Overall & 55.10  & \cellcolor{yellow!40}69.04  & 79.22  & \cellcolor{yellow!40}\textbf{92.43} & 67.16 & \cellcolor{yellow!40}\textbf{80.73} \\
        
        \midrule
        \multirow{3}{*}{Qwen 32B} 
        & Set-G  & 78.15  & \cellcolor{yellow!40}78.95  & \cellcolor{yellow!40}58.71  & 58.45 & 68.43 & \cellcolor{yellow!40}68.70 \\
        & Set-N  & 29.89  & \cellcolor{yellow!40}\textbf{94.24} & \cellcolor{yellow!40}78.95  & 61.39 & 54.42 & \cellcolor{yellow!40}\textbf{77.82} \\
        & Overall & 54.02  & \cellcolor{yellow!40}86.60  & \cellcolor{yellow!40}68.83  & 59.92 & 61.43 & \cellcolor{yellow!40}73.26 \\

        \midrule
        \multirow{3}{*}{Qwen 72B} 
        & Set-G  & 75.20  & \cellcolor{yellow!40}\textbf{91.29}  & 12.06  & \cellcolor{yellow!40}65.01 & 43.63 & \cellcolor{yellow!40}78.15 \\
        & Set-N  & 42.90  & \cellcolor{yellow!40}72.39 & 57.37  & \cellcolor{yellow!40}82.04 & 50.14 & \cellcolor{yellow!40}77.22 \\
        & Overall & 59.05  & \cellcolor{yellow!40}81.84  & 34.72  & \cellcolor{yellow!40}73.53 & 46.88 & \cellcolor{yellow!40}77.68 \\
        
        \midrule
        \multirow{3}{*}{Mistral Small} 
        & Set-G  & \cellcolor{yellow!40} 76.14  & 65.68  & 23.86  & \cellcolor{yellow!40}39.95 & 50.00 & \cellcolor{yellow!40}52.82 \\
        & Set-N  & 31.10  & \cellcolor{yellow!40}92.63  & \cellcolor{yellow!40}69.97  & 55.23 & 50.54 & \cellcolor{yellow!40}73.93 \\
        & Overall & 53.62  & \cellcolor{yellow!40}79.16  & 23.86  & \cellcolor{yellow!40}47.59 & 38.74 & \cellcolor{yellow!40}63.37 \\

        \midrule
        \multirow{3}{*}{DS Qwen 32B} 
        & Set-G  & 43.57  & \cellcolor{yellow!40}84.05  & 32.04  & \cellcolor{yellow!40}51.34 & 37.81 & \cellcolor{yellow!40}67.70 \\
        & Set-N  & 21.98  & \cellcolor{yellow!40}89.95  & \cellcolor{yellow!40}50.94  & 50.00 & 36.46 & \cellcolor{yellow!40}69.98 \\
        & Overall &32.78  & \cellcolor{yellow!40}\textbf{87.00}  & 41.49  & \cellcolor{yellow!40}50.67 & 37.13 &\cellcolor{yellow!40}68.84\\
        \bottomrule

    \end{tabular}
    \caption{
    \ap{Accuracy of all models in}
    \textit{source-target} English~→~Spanish GNT evaluation on mGeNTE source-reference pairs. The best-performing settings for each data split are in bold. The best performing strategy per model and data split is highlighted.}
    \label{tab:cross-results-es}
\end{table*}

\end{document}